\documentclass[11pt]{article}

\usepackage{anthology}

\usepackage{times}
\usepackage{array}
\usepackage{bm}
\usepackage{latexsym}
\usepackage{mathrsfs}
\usepackage{amsfonts} 
\usepackage{amssymb}
\usepackage{amsmath}
\usepackage{multirow}
\usepackage{bbding}
\usepackage{booktabs}
\usepackage{tabularx}
\usepackage{color}
\usepackage{graphicx}
\graphicspath{{figure/} }
\DeclareGraphicsExtensions{.pdf,.png}

\usepackage[T1]{fontenc}

\usepackage[utf8]{inputenc}

\usepackage{microtype}

\newcommand{\revise}[1]{\textcolor{black}{#1}} 

\title{Multimodal Emotion-Cause Pair Extraction 
in Conversations}

\author
{
  Fanfan Wang\thanks{\hspace{0.3em} Equal contribution.}
  \hspace{10pt} Zixiang Ding$^{*}$
  \hspace{10pt} Rui Xia\thanks{\hspace{0.3em} Corresponding author.}
  \hspace{10pt} Zhaoyu Li
  \hspace{10pt} Jianfei Yu\\
  School of Computer Science and Engineering,\\
  Nanjing University of Science and Technology, China\\
  
  \texttt{\{ffwang, dingzixiang, rxia, zyli, jfyu\}@njust.edu.cn} \\
}

\begin{document}
\maketitle
\begin{abstract}
Emotion cause analysis has received considerable attention in recent years. 
Previous studies primarily focused on emotion cause extraction from texts in news articles or microblogs.
It is also interesting to discover emotions and their causes in conversations. 
As conversation in its natural form is multimodal, 
a large number of studies have been carried out on multimodal emotion recognition in conversations, but there is still a lack of work on multimodal emotion cause analysis.
In this work, we introduce a new task named Multimodal Emotion-Cause Pair Extraction in Conversations, aiming to jointly extract emotions and their associated causes from conversations reflected in multiple modalities 
(text, audio and video).
We accordingly construct a multimodal conversational emotion cause dataset, Emotion-Cause-in-Friends, which contains 9,272 multimodal emotion-cause pairs annotated on 13,509 utterances in the sitcom \textit{Friends}. 
We finally benchmark the task by establishing a baseline system that incorporates multimodal features for emotion-cause pair extraction. Preliminary experimental results demonstrate the potential of multimodal information fusion for discovering both emotions and causes in conversations.
\end{abstract}

\section{Introduction}

In the field of textual emotion analysis, previous research mostly focused on emotion recognition. In recent years, emotion cause analysis, a new task which aimed at extracting potential causes given the emotions \citep{lee2010text,chen2010emotion,gui2016emotion} or jointly extracting emotions and the corresponding causes in pairs \citep{xia2019emotion,ding2020ecpe,wei2020effective,fan2020transition}, has received much attention. These studies were normally carried out based on news articles or microblogs. \citet{poria2021recognizing} further introduced an interesting task to recognize emotion cause in textual dialogues.

However, conversation in its natural form is multimodal. Multimodality is especially important for discovering both emotions and their causes in conversations. For example, we do not only rely on the speaker’s voice intonation and facial expressions to perceive his emotions, but also depend on some auditory and visual scenes to speculate the potential causes that trigger the speakers' emotions beyond text. 
Although a large number of studies have explored multimodal emotion analysis in conversations~\citep{busso2008iemocap,mckeown2012semaine,poria2019meld}, to our knowledge, at present there is still a lack of research on multimodal emotion cause analysis in conversations.

\begin{figure*}[htb]
    \centering
    \includegraphics[width=\textwidth]{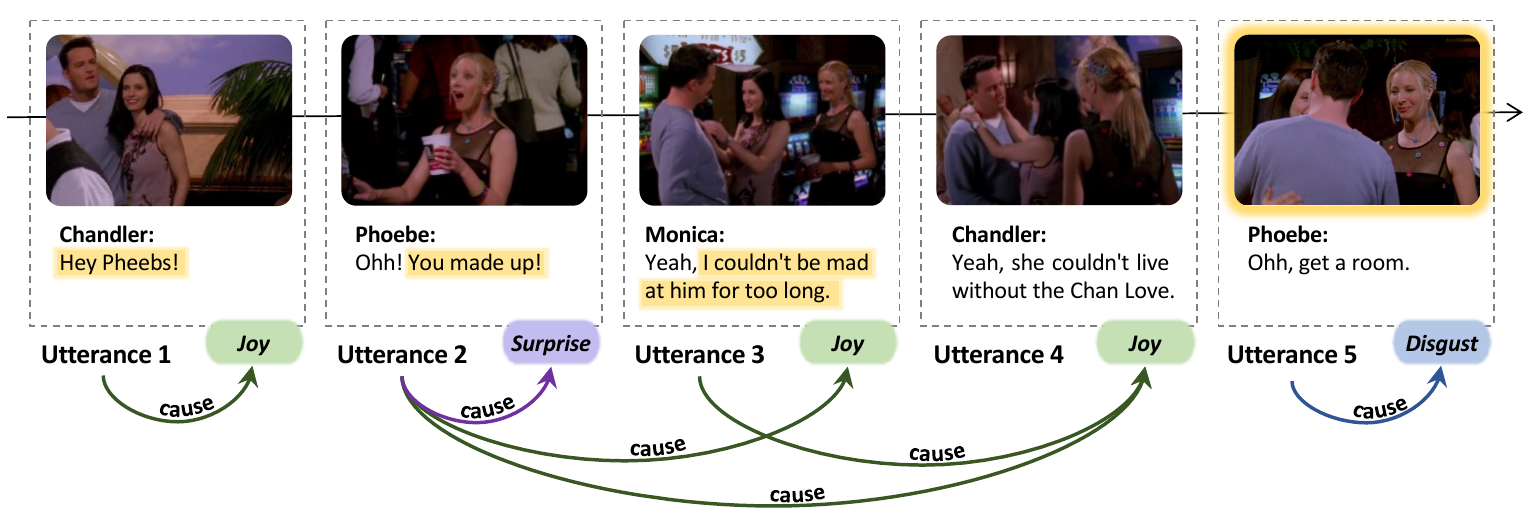}
    \caption{An example of the annotated conversation in our ECF dataset. Each arc points from the cause utterance to the emotion it triggers. We have highlighted the cause evidence which is mainly contained in a certain modality of the  cause utterance.
    }
    \label{fig:example}
\end{figure*}

In this work, we introduce a new task named Multimodal Emotion-Cause Pair Extraction in Conversations (MC-ECPE), with the goal to extract all potential pairs of emotions and their corresponding causes from a conversation in consideration of three modalities 
(text, audio and video).
We accordingly construct a multimodal emotion cause dataset, Emotion-Cause-in-Friends (ECF), by using the sitcom \textit{Friends} as the source. The ECF dataset contains 1,344 conversations and 13,509 utterances\footnote{An utterance is a unit of speech divided by the speaker's breath or pause.}, where 9,272 emotion-cause pairs are annotated, covering three modalities.

Figure \ref{fig:example} displays a real conversation in the ECF dataset, where Chandler and his girlfriend Monica walked into the casino, hugging each other (they had a quarrel earlier but made up soon), and then started a conversation with Phoebe. 
In Utterance 1 ($U_1$ for short), Chandler said hello to Phoebe with a \textit{Joy} emotion (the cause is greeting).
Phoebe's \textit{Surprise} emotion in $U_2$ is caused by the event that Chandler and Monica had made up (reflected by the textual modality in $U_2$). This is also the cause of Monica's \textit{Joy} emotion in $U_3$ and Chandler's \textit{Joy} emotion in $U_4$. Chandler's \textit{Joy} emotion in $U_4$ has another cause - Monica's opinion in $U_3$ (``\textit{I couldn't be mad at him for too long}''). The cause for Phoebe's \textit{Disgust} in $U_5$ is the event that Monica and Chandler were kissing in front of her. This is not explicitly expressed by the textual modality, but is mainly reflected 
in the visual modality of $U_5$. 
For this example, it is expected to extract a set of six utterance-level emotion-cause pairs: 
\{$U_1\text{-}U_1$, $U_2\text{-}U_2$, $U_3\text{-}U_2$, $U_4\text{-}U_2$, $U_4\text{-}U_3$, $U_5\text{-}U_5$\}.

We finally benchmark the MC-ECPE task by establishing a baseline system adapted from a representative textual ECPE approach. We incorporate multimodal features for utterance representation, extract emotion utterances and cause utterances respectively, and finally construct emotion-cause pairs. The experimental results demonstrate the effect of multimodal information fusion for discovering both emotions and causes in conversations.

\begin{table*}[ht]
\centering
\small
{
\begin{tabular}{|l|c|c|c|l|}
\toprule[1pt]
\multicolumn{1}{|l|}{\textbf{Dataset}} & 
\multicolumn{1}{c|}{\textbf{ Modality}} & 
\multicolumn{1}{c|}{\textbf{ Cause}} & 
\multicolumn{1}{c|}{\textbf{ Scene}} & 
\multicolumn{1}{l|}{\textbf{ \# Instances}} \\ 
\midrule[1pt]
Emotion-Stimulus  \citep{ghazi2015detecting} & T & \Checkmark & -- & 2,414 sentences \\
ECE Corpus \citep{gui2016event} & T & \Checkmark & News & 2,105 documents \\
NTCIR-13-ECA  \citep{gao2017overview} & T & \Checkmark & Fiction & 2,403 documents \\
Weibo-Emotion \citep{cheng2017emotion} & T & \Checkmark & Microblog & 7,000 posts \\
REMAN \citep{kim2018feels} & T & \Checkmark & Fiction & 1,720 documents \\
GoodNewsEveryone \citep{bostan2020goodnewseveryone} & T & \Checkmark & News & 5,000 sentences \\
\midrule[0.5pt]
IEMOCAP \citep{busso2008iemocap} & T,A,V & \XSolidBrush & Conversation & 7,433 utterances \\
DailyDialog \citep{li2017dailydialog} & T & \XSolidBrush & Conversation & 102,979 utterances \\
EmotionLines \citep{hsu2018emotionlines} & T & \XSolidBrush & Conversation & 14,503 utterances \\
SEMAINE \citep{mckeown2012semaine} & T,A,V & \XSolidBrush & Conversation & 5,798 utterances \\
EmoContext \citep{chatterjee2019understanding} & T & \XSolidBrush & Conversation & 115,272 utterances \\
MELD \citep{poria2019meld} & T,A,V & \XSolidBrush & Conversation & 13,708 utterances \\
MELSD \citep{firdaus2020meisd} & T,A,V & \XSolidBrush & Conversation & 20,000 utterances \\
\midrule[0.5pt]
RECCON-IE \citep{poria2021recognizing} & T & \Checkmark & Conversation & 665 utterances \\
RECCON-DD \citep{poria2021recognizing} & T & \Checkmark & Conversation & 11,104 utterances \\
\textbf{ECF }(ours)  & T,A,V & \Checkmark & Conversation & 13,509 utterances \\ 
\bottomrule[1pt]
\end{tabular}%
}
\caption{A summary of datasets for emotion cause analysis and emotion recognition in conversations. T, A, V stand for text, audio and video respectively. }
\label{tab:datasets}
\end{table*}

\section{Related Work}
\label{sec:related work}

\paragraph{Emotion Cause Analysis:} 
Emotion cause extraction (ECE) is a subtask of emotion analysis. It was originally proposed by \citet{lee2010text}, with the goal to extract cause spans of a given emotion in the text. Based on the same task setting, some researchers use rule-based methods \citep{neviarouskaya2013extracting,li2014text,gao2015emotion,gao2015rule,yada2017bootstrap} or machine learning methods \citep{ghazi2015detecting,song2015detecting} to extract emotion causes in their own corpus.

By analyzing the corpus proposed by \citet{lee2010text}, \citet{chen2010emotion} pointed out that clause may be a more suitable unit for cause annotation, and proposed to extract emotion cause at clause granularity. After that, a lot of work based on this task setting appeared \citep{russo2011emocause,gui2014emotion}. Especially, \citet{gui2016event} released an open Chinese emotion cause dataset. This dataset has received extensive attention and become the benchmark dataset for the ECE task. Based on this corpus, many traditional machine learning methods \citep{gui2016event,gui2016emotion,xu2017ensemble} and deep learning methods \citep{gui2017question,li2018co,yu2019multiple,xu2019extracting,ding2019independent,xia2019emotion} were put forward.

However, there are two shortcomings in the ECE task: 1) emotions must be manually annotated before cause extraction, which greatly limits its practical application; 2) the way of annotating emotions first and then extracting causes ignores the fact that emotions and causes are mutually indicative. To solve these problems,
\citet{xia2019emotion} proposed a new task called emotion-cause pair extraction (ECPE), aiming at extracting potential emotions and corresponding causes from documents simultaneously. They further constructed the ECPE dataset based on the benchmark corpus for ECE \citep{gui2016event}. After that, a lot of work on the ECPE task has been put forward to solve the shortcomings of the existing methodology \citep{ding2020ecpe,ding2020end,wei2020effective,fan2020transition}.

The above studies mostly focused on emotion cause analysis in news articles \citep{gui2016event,gao2017overview,bostan2020goodnewseveryone}, microblogs \citep{cheng2017emotion} and fictions \citep{gao2017overview,kim2018feels}.
Recently, \citet{poria2021recognizing} introduced an interesting task of recognizing emotion cause in conversations and constructed a new dataset RECCON for this task.
Considering that conversation itself is multimodal, we further propose to jointly extract emotions and their corresponding causes from conversations based on multiple modalities, and accordingly create a multimodal conversational emotion cause dataset.

\paragraph{Emotion Recognition in Conversations:}
\revise{Although there's a lack of research on multimodal emotion cause analysis, many studies have been carried out on multimodal emotion recognition using textual, acoustic, and visual modalities, especially in conversations \citep{busso2008iemocap,mckeown2012semaine,poria2019meld,hazarika2018icon,jin2020hierarchical}.}

In recent years, due to the increasing amount of open conversation data, the Emotion Recognition in Conversations (ERC) task has received continuous attention in the field of NLP. 
So far, there have been some publicly available datasets for ERC. IEMOCAP \citep{busso2008iemocap} contains multimodal dyadic conversations of ten actors performing the emotional scripts. SEMAINE \citep{mckeown2012semaine} contains multimodal data of robot-human conversations (it does not provide emotion categories, but the attributes of four emotion dimensions). 
The above two datasets are relatively small in scale and do not contain multi-party conversations.
DailyDialog \citep{li2017dailydialog} is a large dataset that contains 
the texts of daily conversations covering 10 topics
, but the neutral utterances in it account for a high proportion. EmoContext \citep{chatterjee2019understanding} has a large total number of utterances, but only contains two-person conversations in plain text, with only three utterances in each conversation. EmotionLines \citep{hsu2018emotionlines} contains two datasets: multi-party conversations from the sitcom \textit{Friends} (Friends) and private chats on Facebook Messenger (EmotionPush) where all the utterances are labeled with emotion categories. \citet{poria2019meld} extended EmotionLines (Friends)
to the multimodal dataset MELD with raw videos, audio segments and transcripts, the size of which is moderate. Recently, \citet{firdaus2020meisd} constructed a large-scale multimodal conversational dataset MEISD from 10 famous TV series, where an utterance may be labeled with multiple emotions along with their corresponding intensities.

\section{Task}
\label{sec:task}

We first clarify the definitions of emotion and cause in our work:

\begin{itemize}

    \item \textbf{Emotion} is a psychological state associated with thought, feeling and behavioral response. In computer science, emotions are often described as discrete emotion categories, such as Ekman’s six basic emotions including \textit{Anger}, \textit{Disgust}, \textit{Fear}, \textit{Joy}, \textit{Sadness} and \textit{Surprise} \citep{1971Universals}. In conversations, emotions are usually annotated at the utterance level \citep{li2017dailydialog, hsu2018emotionlines,poria2019meld}.
    
    \item \textbf{Cause} refers to the explicitly expressed event or argument that is highly linked with the corresponding emotion \citep{lee2010text,chen2010emotion,russo2011emocause}. 
    In this work, we use an utterance to describe an emotion cause. Although we have annotated the textual span if the cause is reflected in the textual modality, we only consider utterance-level emotion and cause extraction in this work, in order to facilitate the representation and fusion of multimodal information. 
    
\end{itemize}

There are two kinds of textual emotion cause analysis task: emotion cause extraction (ECE) \cite{gui2016emotion} and emotion-cause pair extraction (ECPE) \cite{xia2019emotion}. The goal of ECE is to extract the potential causes given the emotion annotation; while ECPE aims to jointly extract the emotions and the corresponding causes in pairs, which solves ECE's shortcoming of emotion annotation dependency and improves the performance of emotion and cause extraction.

Therefore, in this work we directly define the task of multimodal emotion cause analysis under ECPE rather than ECE.
Given a conversation \(D=[U_1,\dots,U_i,\dots,U_{\left|D\right|}] \), 
in which each utterance is represented by the text, audio and video
, i.e., $U_i=[t_i,a_i,v_i]$, 
the goal of MC-ECPE is to extract a set of emotion-cause pairs:
\begin{equation}
\mathcal{P}=\left\{\dots, U^e\text{-}U^c,\dots \right\},
\end{equation}
where \(U^e \) denotes an emotion utterance and \(U^c \) is the corresponding cause utterance.

\section{Dataset}

\subsection{Dataset Source}
The conversations in sitcoms usually contain more emotions than other TV series and movies.
\citet{hsu2018emotionlines} constructed the EmotionLines dataset from the scripts of the popular sitcom \textit{Friends} for the ERC task.
\citet{poria2019meld} extended EmotionLines to a multimodal dataset MELD, by extracting the audio-visual clip from the source episode, and re-annotating each utterance with emotion labels.

We find that sitcoms also contain rich emotion causes, therefore we choose MELD\footnote{\href{https://github.com/declare-lab/MELD}{MELD} is licensed under GNU General Public License v3.0. } as the data source and further annotate the corresponding causes for the given emotion annotations.
We drop a few conversations where three modalities are completely inconsistent in timestamps.

\begin{figure}
    \centering
    \includegraphics[width=0.48\textwidth]{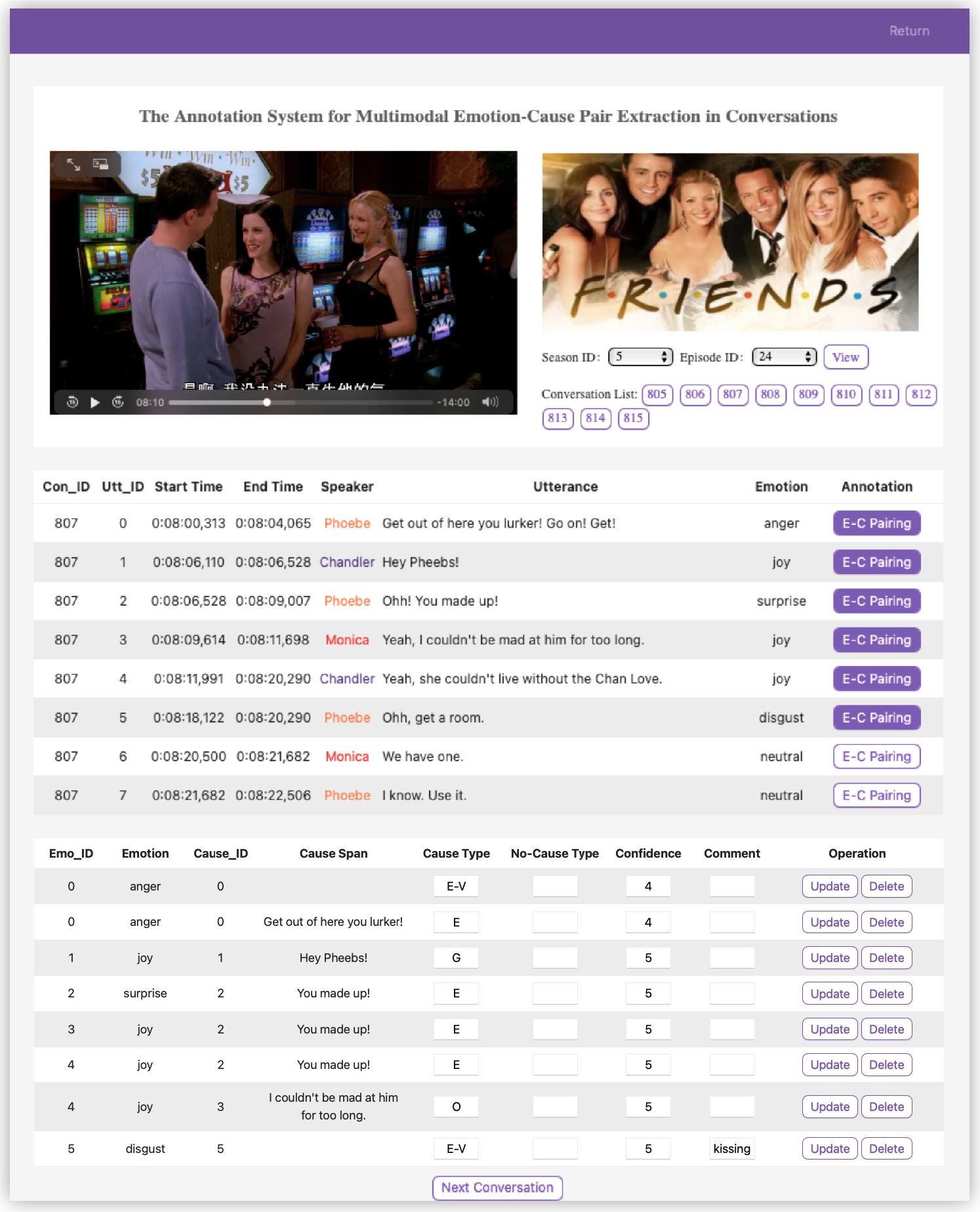}
    \caption{The interface of our developed annotation toolkit.}
    \label{fig:tool}
\end{figure}

\begin{table*}[ht]
\small
\centering
\resizebox{\textwidth}{!}
{%
\begin{tabular}{|m{0.1\textwidth}|m{0.25\textwidth}|c|c|m{0.48\textwidth}|}
\toprule[1pt]
\multicolumn{1}{|l|}{\textbf{Type}} & 
\multicolumn{1}{l|}{\textbf{Explanation}} & 
\multicolumn{1}{c|}{\textbf{Modality}} & 
\multicolumn{1}{c|}{\textbf{\%}} & 
\multicolumn{1}{l|}{\textbf{Example}} \\ 
\midrule[1pt]
\multirow{11}{0.1\textwidth}{Event} & \multirow{11}{0.25\textwidth}{Something that happens in a particular situation, which is normally a fact.} & \multirow{2}{0.05\textwidth}{\centering T} & \multirow{2}{0.07\textwidth}{\centering 60.14\%} & {[}U1{]} Phoebe: Ohh! {\color{blue} You made up}!  ({\color{red} \textit{Surprise}}) \\
 &  &  &  & \textbf{Emotion-Cause Pair:} (U1, U1) \\ \cline{3-5}
 &  & \multirow{5}{0.05\textwidth}{\centering A} & \multirow{5}{0.07\textwidth}{\centering 0.60\%}  & [U1] Chandler: {\color{blue} What is wrong with Emma?} ({\color{red} \textit{Sadness}}) \\ 
 &  &  &  & \multirow{2}{0.48\textwidth}{[U2] Monica: Oh she misunderstood, she thought she was moving to Tulsa. (\textit{Neutral})} \\ 
 &  &  &  &  \\ 
 &  &  &  & \textbf{Emotion-Cause Pair:} (U1, U1) \\ 
 &  &  &  & *Note: Chandler heard Emma crying. \\ \cline{3-5}
 &  & \multirow{4}{0.05\textwidth}{\centering V} & \multirow{4}{0.07\textwidth}{\centering 7.56\%}  & [U1] Phoebe: Ohh, get a room. ({\color{red} \textit{Disgust}}) \\ 
 &  &  &  & \textbf{Emotion-Cause Pair:} (U1, U1) \\  
 &  &  &  & \multirow{2}{0.48\textwidth}{*Note: In the video, Monica and Chandler were kissing in front of Phoebe, as shown in Figure \ref{fig:example}.} \\
 &  &  &  &   \\
\midrule[0.5pt]
\multirow{5}{0.1\textwidth}{Opinion} & \multirow{5}{0.25\textwidth}{Someone's feelings or thoughts about people or things rather than a fact.} & \multirow{5}{0.05\textwidth}{\centering T}  & \multirow{5}{0.07\textwidth}{\centering 25.11\%} & \multirow{2}{0.48\textwidth}{{[}U1{]} Monica: Yeah, {\color{blue} I couldn't be mad at him for too long}. (\textit{Joy})} \\
 &  &  &  &   \\
 &  &  &  & \multirow{2}{0.48\textwidth}{{[}U2{]} Chandler: Yeah, she couldn't live without the Chan Love. ({\color{red} \textit{Joy}})} \\
 &  &  &  &  \\
 &  &  &  & \textbf{Emotion-Cause Pair:} (U2, U1) \\
\midrule[0.5pt]
\multirow{4}{0.1\textwidth}{Emotional Influence} & \multirow{4}{0.25\textwidth}{The speaker's emotion is induced by the counterpart's emotion.} & \multirow{4}{0.05\textwidth}{\centering T,A,V} & \multirow{4}{0.07\textwidth}{\centering 3.74\%} & \multirow{2}{0.48\textwidth}{{[}U1{]} Joey: Fine, you want to get the birds, get the birds! (\textit{Anger})} \\
 &  &  &  &  \\
 &  &  &  &  {[}U2{]} Chandler: Not like that, I won't! ({\color{red} \textit{Sadness}})\\ 
 &  &  &  & \textbf{Emotion-Cause Pair:} (U2, U1) \\
\midrule[0.5pt]
\multirow{2}{0.1\textwidth}{Greeting} & \multirow{2}{0.25\textwidth}{People tend to be happy when they meet and greet each other.
} & \multirow{2}{0.05\textwidth}{\centering T,V} & \multirow{2}{0.07\textwidth}{\centering 2.85\%} & [U1] Chandler: {\color{blue} Hey Pheebs!} ({\color{red} \textit{Joy}}) \\
 &  &  &  & \textbf{Emotion-Cause Pair:} (U1, U1)  \\
\bottomrule[1pt]
\end{tabular}%
}
\caption{A summary of emotion cause types. In the example, the emotion and textual cause span are colored in red and blue accordingly.}
\label{tab:cause type}
\end{table*}

\subsection{Annotation Procedure}
Given a multimodal conversation, for the emotion (one of Ekman's six basic emotions)
labeled on each utterance, the annotator should annotate the utterances containing corresponding causes, label the types of causes, and mark the textual cause spans if the causes are explicitly expressed in the textual modality.

We first develop detailed annotation instructions and then employ three annotators who have reasonable knowledge of our task to annotate the entire dataset independently. 
After annotation, we determine the cause utterances by majority voting
and take the largest boundary (i.e., the union of the spans) as the gold annotation of textual cause span,
similar as \citep{gui2016event,bostan2020goodnewseveryone}. If there are further disagreements, another
expert is invited for the final decision.

To improve the annotation efficiency, we furthermore develop a multimodal emotion cause annotation toolkit\footnote{We will release this toolkit as open-source software, together with the ECF dataset, to facilitate subsequent research and annotations for this task.}. It is a general toolkit for multimodal annotation in conversations, with the functions of multimodal signal alignment, quick emotion-cause pair selection, multiple users and tasks management, distributable deployment, etc. 
Figure \ref{fig:tool} displays the interface of the toolkit.

\begin{table}
\centering
\small
{%
\begin{tabular}{|c|cccc|}
\toprule[1pt]
\textbf{Annotator Pair} & A\&B  & A\&C & B\&C & Avg. \\
\midrule[1pt]
\textbf{Cohen's Kappa}   & 0.6348  & 0.6595  & 0.6483  & 0.6475  \\
\textbf{Fleiss' Kappa}   &   \multicolumn{4}{c|}{0.6044}      \\

\bottomrule[1pt]
\end{tabular}
}
\caption{The inter-annotator agreement for utterance-level emotion cause annotations. A, B, C represent the three annotators respectively. 
}
\label{tab:IAA}
\end{table}

\subsection{Annotation Quality Assessment}
To evaluate the quality of annotation, we measure the inter-annotator agreement on the full set of annotations, based on 
Cohen’s Kappa and Fleiss’ Kappa.
Cohen’s Kappa is used to measure the consistency of any two annotators~\cite{cohen1960coefficient}, while Fleiss’ Kappa is used to measure the overall annotation consistency among three annotators~\cite{mchugh2012interrater}.
The agreement scores are reported in Table \ref{tab:IAA}. 

It can be seen that the Kappa coefficients are all higher than 0.6, which indicates a substantial agreement between three annotators \citep{landis1977measurement}.

\begin{table}
\centering
\small
{%
\begin{tabular}{|l|c|}
\toprule[1pt]
\textbf{Items} & \textbf{Number} \\
\midrule[1pt]
Conversations & 1,344 \\
Utterances & 13,509 \\
Emotion (utterances) & 7,528 \\
Emotion (utterances) with cause & 6,876 \\
Emotion-cause (utterance) pairs & 9,272 \\
\bottomrule[1pt]
\end{tabular}%
}
\caption{Basic statistic of our ECF dataset.}
\label{tab:statistic}
\end{table}

\subsection{Dataset Statistic and Analysis}
\label{subsec:data statis}

\paragraph{Overall Statistics:}
As shown in Table \ref{tab:statistic}, the ECF dataset contains 1,344 conversations and 13,509 utterances from three modalities, where 7,528 emotion utterances and 9,272 emotion-cause pairs have been annotated.
In other words, about 55.73\% of the utterances are annotated with one of the six basic emotions, and 91.34\% of the emotions are annotated with the corresponding causes in our dataset. The number of pairs is larger than 6,876, which indicates that one emotion may be triggered by multiple causes in different utterances.

In Table \ref{tab:datasets}, we compare our ECF dataset with the related datasets in the field of emotion cause analysis and emotion recognition in conversations, in terms of modality, scene, and size.

\begin{figure*}[ht]
    \centering
    \includegraphics[width=\textwidth]{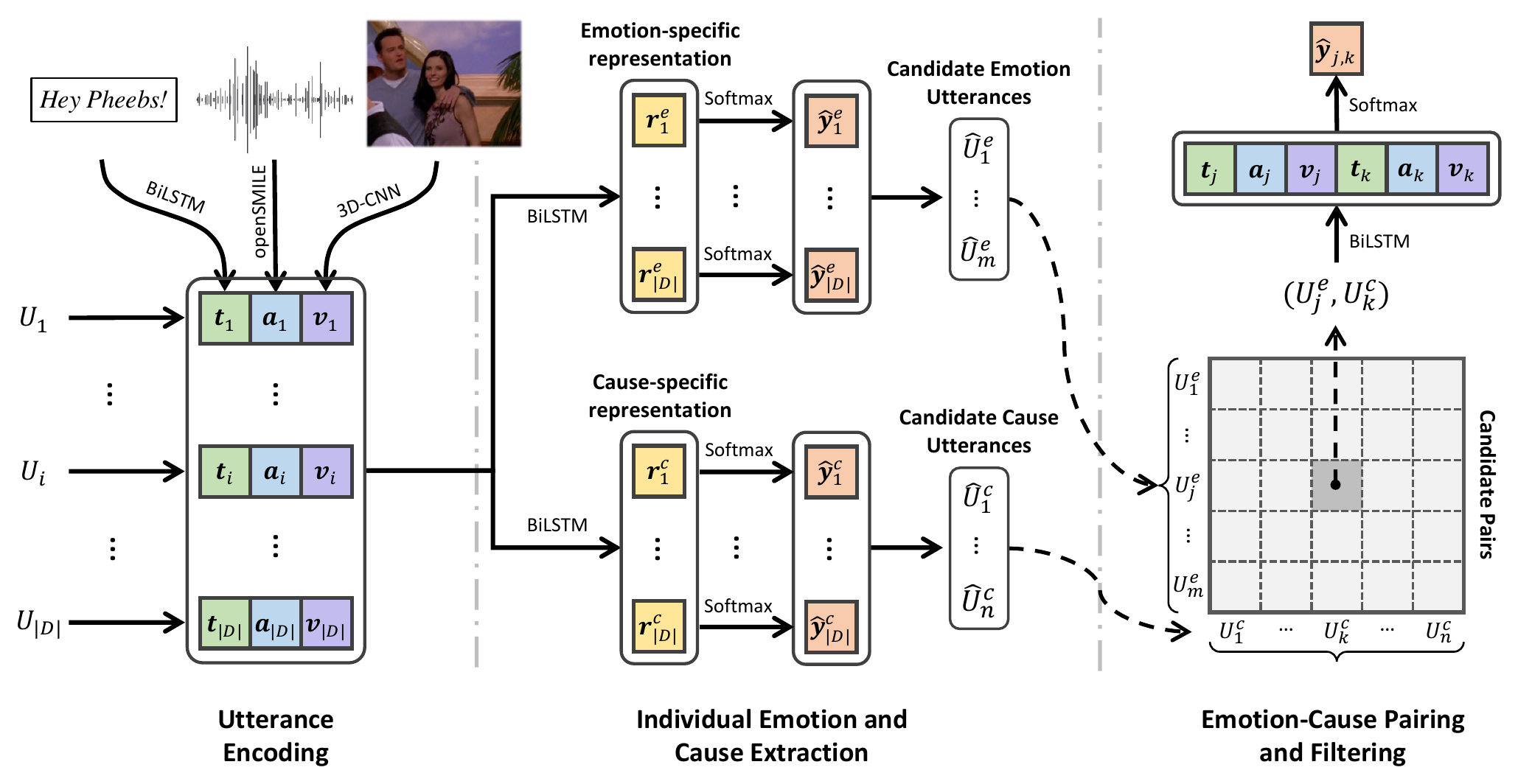}
    \caption{Overview of the baseline system MC-ECPE-2steps.}
    \label{fig:model}
\end{figure*}

\begin{figure}
    \centering
    \includegraphics[width=0.48\textwidth]{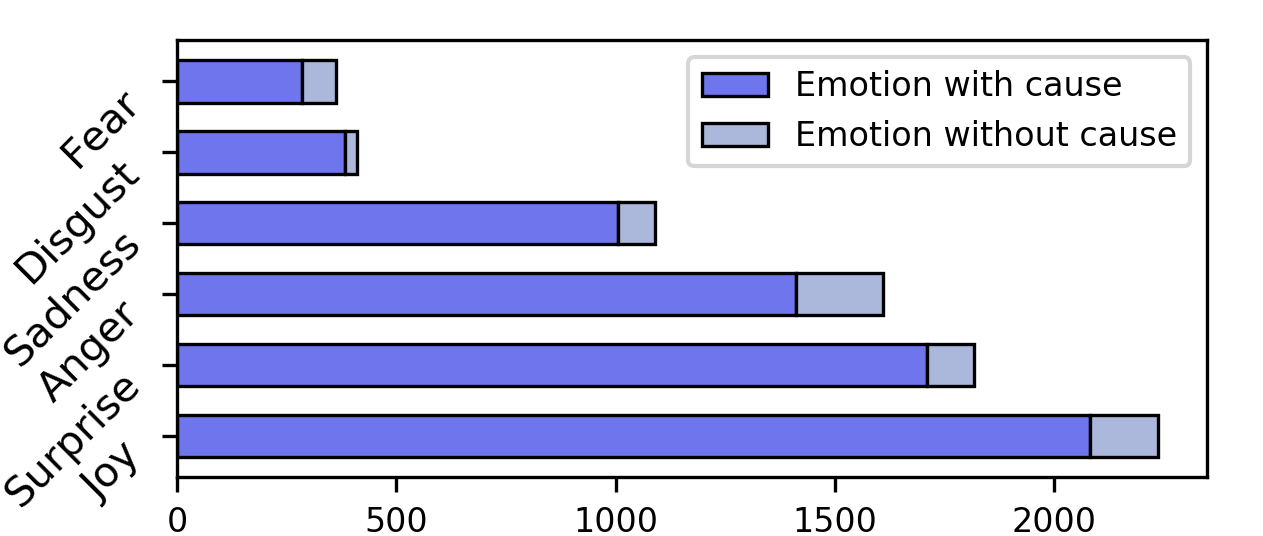}
    \caption{The distribution of emotions (with/without cause) in different categories.}
    \label{fig:distribution}
\end{figure}

\paragraph{Emotion/Cause Distribution:}
For each emotion category, the proportion of emotion utterances annotated with causes is shown in Figure \ref{fig:distribution}. It can be seen that the distribution of emotion categories is unbalanced, and the proportion of emotion having causes varies slightly with the emotion category.

\paragraph{Types of Emotion Causes:}

In Table \ref{tab:cause type}, 
we furthermore summarized the emotion causes in our dataset into four types.

\begin{itemize}
    \item \textbf{Event: } The speaker's emotion is caused by something that happens in a particular situation, which is normally a fact. This type of cause may be reflected in the three modalities.
    \item \textbf{Opinion:} The speaker's emotion is triggered by someone's feelings or thoughts about people or things rather than a fact. Such causes are only expressed in texts.
    \item \textbf{Emotional Influence:} The speaker's emotion is sometimes induced by the counterpart's emotion. 
    This type of cause is normally embodied in three modalities jointly. 
    \item \textbf{Greeting:} As an act of giving a sign of welcome or recognition in communication, the greeting is a cause for the \textit{Joy} emotion in daily conversations. It can be reflected in both textual and visual modalities.
\end{itemize}

We can see that ``Event'' covers the largest percentage of emotion causes (68.30\%), followed by ``Opinion'' (25.11\%), which suggests that most of the emotions in conversations are triggered by specific events or subjective opinions. 

It is also worth noting that 8\% of the emotion causes in our dataset are the events mainly reflected in the acoustic or visual modalities.
For example, Phoebe was disgust because Monica and Chandler were kissing in front of her in the scene shown in Figure \ref{fig:example}, we cannot speculate on such causes only based on the textual content obviously.

\section{Baseline}
\label{sec:method}
In this section, we benchmark our MC-ECPE task by proposing a baseline system named MC-ECPE-2steps, which is adapted from a representative ECPE-2steps \citep{xia2019emotion} approach for ECPE in news articles.

\subsection{Main Framework}
\label{subsec:baseline}
The main framework of MC-ECPE-2steps is shown in Figure \ref{fig:model}.

Step 1 aims to extract a set of emotion utterances and a set of cause utterances individually via multi-task learning.
We first obtain the independent utterance representations $\bm{u}_{i}$ through word-level encoder and then feed them into two utterance-level encoders. The hidden states $\bm{r}_i^e$ and $\bm{r}_i^c$, which can be viewed as the emotion-specific representation and cause-specific representation of utterance $U_i$, are used to perform emotion prediction and cause prediction respectively. 

Step 2 performs emotion-cause pairing and filtering. We combine all predicted emotions and causes into pairs, obtain the pair representation through BiLSTM and attention mechanism,  and finally filter out the pairs that do not contain a causal relationship via a feed-forward neural network.

\subsection{Multimodal Features}
\label{subsec:feature}

Since the emotions and causes in this paper are defined on the multimodal utterances, we further extract the features from three modalities 
and then concatenate them to obtain the independent multimodal representation of each utterance, i.e., $\bm{u}_i=[\bm{t}_{i},\bm{a}_i,\bm{v}_i]$.

\paragraph{Text:}
We initialize each token with pre-trained 300-dimensional GloVe vectors \citep{pennington2014glove}, feed them into a BiLSTM and then obtain the textual features of each utterance  \(\bm{t}_i \) after an attention mechanism.

\paragraph{Audio:}
We adopt the 1611-dimensional acoustic features \(\bm{a}_i \) extracted by \citet{poria2019meld} using openSMILE in their MELD dataset.

\paragraph{Video:}
We apply 3D-CNN \citep{ji20123d}
to extract the 128-dimensional global scene features \(\bm{v}_i \) from the video of each utterance.

\section{Experiments}

\subsection{Settings and Metrics}
The maximum number of utterances in each conversation and the maximum number of words in each utterance are both set to 35. The dimensions of word embedding and relative position are set to 300 and 50, respectively. The hidden dimension of BiLSTM is set to 100. All models are trained based on the Adam optimizer with a batch size of 32 and a learning rate of 0.005. The dropout ratio is set to 0.5, and the weight of $L_2$-norm regularization is set to 1e-5. 

We divide the dataset into training, validation and testing sets in a ratio of 7:1:2 at the conversation level. In order to obtain statistically credible results,  we repeat all the experiments 20 times and report the average results. 
The precision, recall and $\rm F_{1}$ score defined in \citet{xia2019emotion} are used as the evaluation metrics for the MC-ECPE task.

\begin{table}
\centering
\small
{%
\begin{tabular}{|l|ccc|}
\toprule[1pt]
\textbf{Approach} & \textbf{P} & \textbf{R} & \textbf{F$\rm _{1}$} \\
\midrule[1pt]
$\textrm{E}_{\textrm{Pred}}+\textrm{C}_{\textrm{Multi-Bernoulli}}$ & 0.3662 & 0.2024 & 0.2605  \\
$\textrm{E}_{\textrm{True}}+\textrm{C}_{\textrm{Multi-Bernoulli}}$  & 0.4940 & 0.2522 & 0.3339  \\
$\textrm{E}_{\textrm{Pred}}+\textrm{C}_{\textrm{Multinomial}}$ & 0.3658 & 0.2024 & 0.2604  \\
$\textrm{E}_{\textrm{True}}+\textrm{C}_{\textrm{Multinomial}}$ & 0.4933 & 0.2518 &  0.3334  \\
\midrule[0.5pt]
MC-ECPE-2steps & 0.4943 & 0.5376 & \textbf{0.5132}  \\
\; $\rm - \textrm{Audio}$ & 0.5391 & 0.4778 & 0.5045  \\
\; $\rm - \textrm{Video}$ & 0.4989 & 0.5289 & 0.5116  \\
\; $\rm - \textrm{Audio}-\textrm{Video}$ & 0.5565 & 0.4465 & 0.4942   \\

 \bottomrule[1pt]
\end{tabular}%
}
\caption{Experimental results on the MC-ECPE task. 
}
\label{tab:task1}
\end{table}

\begin{table*}
\centering
\small
{%
\begin{tabular}{|l|cccccc|cc|}
\toprule[1pt]
\textbf{Approach} & \textbf{Anger} & \textbf{Disgust} & \textbf{Fear} & \textbf{Joy} & \textbf{Sadness} & \textbf{Surprise} & \textbf{w-avg. 6} & \textbf{w-avg. 4} \\
\midrule[1pt]
$\textrm{E}_{\textrm{Pred}}+\textrm{C}_{\textrm{Multi-Bernoulli}}$ & 0.1352 & 0.0302 & 0.0146 & 0.1937 & 0.0812 & 0.1905 & 0.1457 & 0.1567 \\
$\textrm{E}_{\textrm{True}}+\textrm{C}_{\textrm{Multi-Bernoulli}}$  & 0.1415 & 0.0318 & 0.0145 & 0.2091 & 0.0885 & 0.2013 & 0.1552 & 0.1670  \\
$\textrm{E}_{\textrm{Pred}}+\textrm{C}_{\textrm{Multinomial}}$  & 0.1346 & 0.0302 & 0.0147 & 0.1928 & 0.0812 & 0.1907 & 0.1453 & 0.1563 \\
$\textrm{E}_{\textrm{True}}+\textrm{C}_{\textrm{Multinomial}}$  & 0.1416 & 0.0322 & 0.0153 & 0.2082 & 0.0878 & 0.2017 & 0.1550 & 0.1667 \\
\midrule[0.5pt]
MC-ECPE-2steps & \textbf{0.2837} & 0.0466 & 0.0260 & \textbf{0.3673} & \textbf{0.1820} & 0.4211 & \textbf{0.3000} & \textbf{0.3237} \\
\; $\rm - \textrm{Audio}$  & 0.2400 & \textbf{0.0630} & \textbf{0.0268} & 0.3511 & 0.1511 & 0.4043 & 0.2764 & 0.2970 \\
\; $\rm - \textrm{Video}$ &  0.2837 & 0.0509 & 0.0262 & 0.3661 & 0.1781 & \textbf{0.4213} & 0.2993 & 0.3227 \\
\; $\rm - \textrm{Audio} - \textrm{Video}$ &  0.2233 & 0.0522 & 0.0175 & 0.3417 & 0.1216 & 0.3982 & 0.2625 & 0.2827   \\
\bottomrule[1pt]
\end{tabular}%
}
\caption{Experimental results on the task of MC-ECPE with emotion category.
*Note: ``w-avg. 4'' denotes the weighted-average $\rm F_{1}$ score of the main four emotions except \textit{Disgust} and \textit{Fear}.}
\label{tab:task2}
\end{table*}

\subsection{Overall Performance}
\label{subsec:exp_overall}

In addition to MC-ECPE-steps, we further design four simple statistical methods for comparison, based on the observation that
there is a certain trend in the relative positions between emotion utterances and cause utterances (i.e., most cause utterances are either the emotion utterances themselves or are immediately before their corresponding emotion utterances). 
In the training phase, we separately train an emotion classifier, and calculate the prior probability distribution of relative positions between the cause utterances and their corresponding emotion utterances.
In the testing phase, we first obtain emotion utterances in two alternative ways: 1) emotion prediction ($\textrm{E}_{\textrm{Pred}}$), which is based on the trained emotion classifier, 2) emotion annotations ($\textrm{E}_{\textrm{True}}$), which are the ground truth labels of emotion prediction. Next, the relative position is assigned to each utterance in the document, which is the position of the current utterance relative to the given emotion utterance (for example, -2, -1, 0, +1, etc.). Then we use the following two strategies to randomly select a cause utterance for each emotion utterance according to the prior probability distribution.

\begin{itemize}
\item  $\textrm{C}_{\textrm{Multi-Bernoulli}}$: We independently carry out a binary decision for each relative position to determine whether its corresponding utterance is the cause utterance. The selection probability of each relative position is calculated from the training set. 

\item  $\textrm{C}_{\textrm{Multinomial}}$: We randomly select a relative position from all relative positions, and its corresponding utterance is the cause utterance. The selection probability of each relative position is calculated from the training set.

\end{itemize}

The experimental results on the MC-ECPE task are reported in table \ref{tab:task1}. 
We can see that the statistical methods perform much poorer than our baseline system MC-ECPE-2steps, which shows that it is not enough to select the emotion cause only based on the relative position.

\subsection{The Effectiveness of Multimodal Features}
\label{subsec:exp_multimodal}

To explore the effectiveness of different multimodal features, 
we conduct extended experiments and also report the results in Table \ref{tab:task1}.

It can be seen that removing the acoustic features or visual features from the baseline system MC-ECPE-2steps ($- \textrm{Audio}$/$- \textrm{Video}$) leads to a decrease in $\rm F_{1}$ score. When both the acoustic features and visual features are removed, the $\rm F_{1}$ score of the system even drops by about 1.9\% 
($\rm  - \textrm{Audio} - \textrm{Video}$). 
Specifically, by integrating multimodal features, the baseline system can predict more causes that are reflected in the auditory and visual scenes, resulting in the great improvement in the recall rate. 
These results illustrate that it's beneficial to introduce multimodal information into the MC-ECPE task.

In addition, we found that the improvement brought by visual features is slightly lower than that brought by acoustic features. This indicates that it is challenging to perceive and understand the complex visual scenes in conversations, hence leaving much room for extra improvement in multimodal feature representation and multimodal fusion.

\subsection{Experiments on MC-ECPE with Emotion Category}
We further conduct experiments on the extended task named ``MC-ECPE with emotion category'' which needs to predict an additional emotion category for each emotion-cause pair.
Specifically, we convert the binary emotion classification to multi-class emotion classification in the first step of MC-ECPE-steps.
We first evaluate the emotion-cause pairs of each emotion category with $\rm F_{1}$ score separately. To evaluate the overall performance, we further calculate a weighted average of $\rm F_{1}$ scores across different emotion categories.
Considering the imbalance of emotion categories in the dataset described in Section \ref{subsec:data statis},  we also report the weighted average $\rm F_{1}$ score  of the main four emotion categories except \textit{Disgust} and \textit{Fear}.

The experimental results on this task are reported in Table \ref{tab:task2}.
An obvious observation is that the performance on \textit{Surprise} is the best, while that on \textit{Fear} is the worst. The performance for different emotion categories significantly varies with the proportion of emotion and cause annotation shown in Figure \ref{fig:distribution}. 
It should be noted that emotion category imbalance is actually an inherent problem in the ERC task \citep{li2017dailydialog,hsu2018emotionlines,poria2019meld}, which is of great challenge and needs to be tackled in future work.

Similar to the conclusion drawn on MC-ECPE, the performance of the baseline system is significantly reduced if not utilizing the acoustic and visual features, which demonstrates that multimodal fusion is also helpful for the task of MC-ECPE with emotion categories.
Although there is much room for further improvement, our model is still effective and feasible. The relatively poor performance on this task indicates that it is more difficult to further predict the emotion categories based on MC-ECPE.

\section{Conclusions and Future Work}

In this work, we introduce a new emotion cause analysis task named Multimodal Emotion-Cause Pair Extraction (MC-ECPE) in Conversations. Secondly, we accordingly construct a multimodal conversational emotion cause dataset, Emotion-Cause-in-Friends (ECF), based on the  American sitcom \textit{Friends}. Finally,  we establish a baseline system and demonstrate the importance of multimodal information for the MC-ECPE task.

MC-ECPE is a challenging task, leaving much room for future improvements. The focus of this work is the introduction of the task and datasets, and we only propose a simple baseline system to benchmark the task. In the future, the following issues are worth exploring in order to further improve the performance of the task: 
\begin{itemize}
\item How to effectively model the impact of speaker relevance on emotion recognition and emotion cause extraction in conversations?
\item How to better perceive and understand the visual scenes to better assist emotion cause reasoning in conversations?
\item How to establish a multimodal conversation representation framework to efficiently align, interact and fuse the information from three modalities?
\end{itemize}

\section*{Acknowledgments}
This work was supported by the Natural Science Foundation of China (No. 62076133 and 62006117), and the Natural Science Foundation of Jiangsu Province for Young Scholars (No. BK20200463) and Distinguished Young Scholars (No. BK20200018).

\bibliography{anthology,custom}
\bibliographystyle{acl_natbib}




\end{document}